\documentclass[journal]{IEEEtran}
 
\usepackage{cite}
\usepackage{amsthm, amssymb, amsmath}
\usepackage{graphicx}
\usepackage{float}
\usepackage{rotating}
\usepackage{subfigure}
\usepackage[justification=centering]{caption}
\usepackage[lined,ruled,linesnumbered,commentsnumbered]{algorithm2e}
\usepackage{cleveref}
\usepackage{booktabs}
\usepackage{amsmath,breqn}
\usepackage{multirow}
\usepackage{amsmath}
\usepackage{pdflscape}
\usepackage{color, colortbl}
\usepackage{threeparttable} 
\usepackage{array}

\newtheorem{definition}{Definition}

\begin{document}

\title{An Online Prediction Approach Based on Incremental Support Vector Machine for Dynamic Multiobjective Optimization}

\author{Dejun~Xu,
        Min~Jiang,~\IEEEmembership{Senior Member,~IEEE,}
        Weizhen~Hu,
        Shaozi~Li,~\IEEEmembership{Senior Member,~IEEE,}
        Renhu~Pan,
        and~Gary G.~Yen,~\IEEEmembership{Fellow,~IEEE}
\thanks{This work was supported by the National Natural Science Foundation of China under Grant 61673328. {\itshape (Corresponding authors: M. Jiang; G. G. Yen.)}}
\thanks{D. Xu, M. Jiang W. Hu and S. Li are with the School of Informatics, Xiamen University, China, Fujian, 361005.}
\thanks{R. Pan is with the Fujian Longking CO., LTD, China, Fujian, 364000.}
\thanks{G. G. Yen is with the School of Electrical and Computer Engineering, Oklahoma State University, USA.}
}

\markboth{}
{Shell \MakeLowercase{\textit{et al.}}: Bare Demo of IEEEtran.cls for IEEE Journals}

\maketitle

\begin{abstract}
Real-world multiobjective optimization problems usually involve conflicting objectives that change over time, which requires the optimization algorithms to quickly track the Pareto optimal front (POF) when the environment changes. In recent years, evolutionary algorithms based on prediction models have been considered promising. However, most existing approaches only make predictions based on the linear correlation between a finite number of optimal solutions in two or three previous environments. These incomplete information extraction strategies may lead to low prediction accuracy in some instances. In this paper, a novel prediction algorithm based on incremental support vector machine (ISVM) is proposed, called ISVM-DMOEA. We treat the solving of dynamic multiobjective optimization problems (DMOPs) as an online learning process, using the continuously obtained optimal solution to update an incremental support vector machine without discarding the solution information at earlier time. ISVM is then used to filter random solutions and generate an initial population for the next moment. To overcome the obstacle of insufficient training samples, a synthetic minority oversampling strategy is implemented before the training of ISVM. The advantage of this approach is that the nonlinear correlation between solutions can be explored online by ISVM, and the information contained in all historical optimal solutions can be exploited to a greater extent. The experimental results and comparison with chosen state-of-the-art algorithms demonstrate that the proposed algorithm can effectively tackle dynamic multiobjective optimization problems.  
\end{abstract}


\IEEEpeerreviewmaketitle

\section{Introduction}

\IEEEPARstart{D}{ynamic} multiobjective optimization problems, which refer to a class of optimization problems involving multiple conflicting objectives and the objective functions or constraints change over time, are very common in urban traffic control, power system scheduling, investment management, data mining and other industrial applications \cite{oyama2010data,barroso2017composition,iqbal2017cross}. For example, in the optimization schedule of multi-reservoirs system for a large-scale hydropower station, engineers need to minimize irrigation water shortage and maximize power generation under the constraints of maintaining water balance and average power generation rate \cite{chang2009multi}. Another convincing example is the emergency supplies allocation after a sudden disaster \cite{bai2016two}: the total distance traveled and allocation time of supply trucks are taken as optimization objectives, with the demand or urgency of each disaster-stricken area and the maximum load of supply trucks comprehensively considered. It is clear that dynamic multiobjective optimization problems are widespread in real-world and play a very important role. However, solving the DMOPs still remains a big challenge due to its constant change in time or environment\cite{cruz2011optimization}. Therefore, the research of dynamic multiobjective optimization algorithms (DMOAs) is of great importance in both theoretical front and practical use. 

In recent years, evolutionary algorithms have been widely used in solving dynamic multiobjective optimization problems \cite{nguyen2012evolutionary}. Evolutionary algorithm usually starts from an initial population, and gradually select the optimal solution in each iteration by specific rules, which in turn efficiently solve some complex problems. In particular, significant progress has been made in a class of prediction-based methods that reuse valuable information from past moments by machine learning and other means. For example, Koo \textit{et al.} \cite{koo2010predictive} proposed a dynamic predictive gradient strategy which estimates the direction and magnitude of the next change based on previous solutions by a weighted average approach. Solutions updated with the predictive gradient will remain in the vicinity of the new Pareto-optimal set and be conducive to population convergence. Zhou \textit{et al.} \cite{zhou2007prediction} presented a method named population prediction strategy (PPS) which maintains a sequence of center points to predict the next center and uses the previous manifold to estimate the next manifold. When changes are detected, PPS can initialize the population by combining the prediction center and estimated manifold. Various models are used in the prediction-based approaches to learn historical knowledge and guide the search, enabling it to respond well to changing environments.

However, room for improvement on the prediction-based evolutionary dynamic multiobjective algorithm still remains. First, most of the existing methods are based on linear prediction model \cite{cao2019evolutionary}. These models can not accurately predict the new solutions if the optimal solutions in DMOPs are nonlinearly correlated at different times. Second, most available methods predict the position of the new Pareto-optimal set based on optimal solutions in previous two or three environments, but realistically only fetching the historical information in the adjacent time may lead to the neglect of some distribution patterns existing in earlier search. Moreover, the accuracy of prediction is highly related to the number of historical optimal solutions. In reality, the optimal solutions obtained by each search are very few compared with the whole decision space. How to extract more information from a limited number of historical optimal solutions remains a challenge.

To address these issues, this paper proposes a novel prediction-based algorithm, called ISVM-DMOEA, which seamlessly integrate several strategies to generate a high-quality population. We believe that there are some implicit correlations between the optimal solutions, from which predictable general patterns can be detected. For a specific problem with a certain regularity, such patterns may exist in all environments experienced. If we can extract the features of the optimal solutions to a greater extent by oversampling method, and constantly assimilate the features via online learning, we can build a more efficient and accurate prediction model for DMOPs. 

The proposed method can be briefly summarized as follows: support vector machine (SVM) is introduced to explore the potential correlations between the optimal solutions at different times, and the features included in the latest optimal solutions are utilized online in an incremental learning process. To further improve the performance of incremental support vector machine (ISVM), we use synthetic minority oversampling technique to deal with the imbalanced data. As the environment changes, a continuously modified ISVM classifier can accurately predict a good initial population for the next moment, which is of great help to the handling of dynamic multiobjective optimization problems.

The contributions of this work are as follows: First, the kernel function in SVM maps the vectors to a high dimensional feature space to construct the classifier, which can handle the possible nonlinear correlation between solutions at different times. Second, the incremental SVM not only can obtain the optimal solution distribution in the new environment online, but also effectively reuse the information contained in all past moments to extract a more comprehensive distribution pattern. Furthermore, the combination of oversampling and incremental SVM overcomes the sample imbalance caused by the small number of optimal solutions. The experimental results show that the algorithm can significantly improve the convergence rate and the quality of solutions, and can be combined with various population-based static optimization algorithms. 

The remainder of the paper is organized as follows: Section \ref{sec:Preliminaries-and-Related-Work} provides the background and some related work of DMOPs. Section \ref{sec:ISVM AND SMOTE} introduces the principles of incremental support vector machine and synthetic minority oversampling technique. Section \ref{sec:Proposed Algorithm} describes the proposed algorithm ISVM-DMOEA in detail. Section  \ref{sec:Experimental Study} presents the experimental study and analysis. Section \ref{sec:Conclusion} concludes the paper with suggestions for future work.

\section{Preliminaries and Related Work}
\label{sec:Preliminaries-and-Related-Work}

\subsection{Dynamic Multiobjective Optimization}
The dynamic multiobjective optimization problem can be defined as: 
\begin{equation}
\left\{\begin{array}{ll}
\textbf{min} & F(x, t)=\left(f_{1}(x, t), f_{2}(x, t), \ldots, f_{M}(x, t)\right) \\
\textbf { s.t. } & x \in \Omega
\end{array}\right.
\end{equation}
where $t$ is the discrete time instants and $x$ is the $D$-dimension decision variable within the decision space $\Omega$.  $F$ refers to the objective vector consists of $M$ time-varying objective functions.

\begin{definition}{\emph{[Pareto Dominance]}}
	At time $t$ , a decision vector $x_{p}$ is said to dominate another vector $x_{q}$, denoted by $x_{p}\succ x_{q}$, if and only if\emph{:}
	\begin{equation}
	\begin{cases}
	\forall i \in \left(1,\ldots,M \right), & f_{i}(x_{p},t)\leq f_{i}(x_{q},t)\\
	\exists i \in \left(1,\ldots,M \right), & f_{i}(x_{p},t)<f_{i}(x_{q},t) .
	\end{cases}
	\end{equation}
\end{definition}

\begin{definition}{\emph{[Dynamic Pareto-optimal Set]}}
      At time  $t$, a solution $x^{*}$ is said to be nondominated (Pareto-optimal) if and only if there is no solution $x$ in the decision space which can dominate $x^{*}$. The Dynamic Pareto-optimal Set (DPOS) is the set of all Pareto-optimal solutions\emph{:}	
	\begin{equation}
\ DPOS(t)=\left\{{x}^{*} \in \Omega \mid \nexists {x} \in \Omega, {x} \succ {x}^{*}\right\}
     \end{equation}
\end{definition}

\begin{definition}{\emph{[Dynamic Pareto-optimal Front]}}
	At time $t$, the Dynamic Pareto-optimal Front includes the corresponding objective vectors of the DPOS\emph{:}
	\begin{equation}
\ DPOF(t)=\left\{{F}\left({x}^{*}, t\right) \mid {x}^{*} \in DPOS(t)\right\}
\end{equation}
\end{definition}

\subsection{Related Work}
Over the years, great progress has been made in the investigation of DMOAs. In general, most existing algorithms can be categorized into three classes: diversity-based approaches, memory-based approaches, and prediction-based approaches.

Diversity-based approaches aim to keep the balance between convergence and diversity. There are two main strategies to enhance the diversity of a population: diversity introduction and diversity maintenance. Diversity introduction can effectively prevent the solutions from trapping in local optima \cite{liu2010sphere}, \cite{azevedo2011generalized}. Deb \textit{et al.} \cite{deb2007dynamic} proposed two variants of NSGA-II for dynamic optimization problems. The first version is called DNSGA-II-A, which introduces randomly generated solutions to replace part of the population; the second version is called DNSGA-II-B, which enhances the diversity by replacing a portion of the population with mutated solutions. Liu \textit{et al.} \cite{liu2020diversity} proposed a method sensitive to change intensity. When environmental change is detected, two strategies are utilized in different situations: an inverse modeling is used for drastic changes, while partially initialization is utilized for mild ones. Ruan \textit{et al.} \cite{ruan2017effect} presented a hybrid diversity algorithm. In an exploitation step, some diverse individuals within the region of the next probable POS are randomly generated.

Some methods based on diversity maintenance were also presented to solve the DMOPs \cite{goh2008competitive}, \cite{li2012achieving}. A steady-state and generational evolutionary algorithm (SGEA) was introduced in \cite{jiang2016steady}, which responds to environmental changes in a steady-state manner. When change occurs, SGEA retains part of outdated solutions with good diversity and predicts some solutions according to the previous environment. These solutions are mixed with random solutions with a certain proportion to create a new population. Shang et al. proposed a class of evolutionary optimization algorithms based on clonal selection \cite{shang2014quantum}, \cite{shang2005clonal}. These algorithms directly use the POS of the current environment as the initial population of the new environment.

Memory-based approaches use additional storage to implicitly or explicitly reserve the solutions in the historical environment, and reuse the stored solutions in the new environment. An adaptive hybrid population management strategy using memory, local search and random strategies was proposed by Azzouz \textit{et al.} in \cite{azzouz2017dynamic}. In this algorithm, the memory size and the number of random solutions to be extracted are dynamically adjusted according to the severity of the change. Xu \textit{et al.} \cite{xu2018memory} presented a memory-enhanced dynamic multiobjective evolutionary algorithm based on Lp decomposition (dMOEA/D-Lp). A subproblem-based bunchy memory scheme is used in dMOEA/D-Lp to store good solutions from past environments and reuse them when necessary. Sahmoud \textit{et al.} \cite{sahmoud2016memory} proposed a hybrid storage strategy integrating memory mechanism within NSGA-II. To improve the ability of NSGA-II to track the non-dominated solutions in dynamic environment, explicit memory is implemented to store the best solutions in each generation. Helbig \textit{et al.} \cite{helbig2011archive} introduced a dynamic vector evaluation particle swarm optimisation (DVEPSO) algorithm and investigated various ways to manage the archive when the environment changes. Chen \textit{et al.} \cite{chen2017dynamic} proposed a two-archive algorithm that dynamically reconstructs two populations (one concerns about convergence and the other concerns about diversity) to solve problems with a time-dependent number of objectives. In general, memory-based mechanism is suitable for DMOPs with periodic changes.

Recently, prediction-based approaches have arisen with an increasing interest among researchers, and a great number of prediction algorithms have been proposed. Essentially, prediction-based approaches use the information of the historical optimal solutions to predict the location of the new POS. 
Muruganantham \textit{et al.} \cite{muruganantham2015evolutionary} proposed a Kalman Filter (KF) based dynamic multiobjective optimization algorithm (MOEA/D-KF). In this method, a linear discrete KF composed of time update equations and measurement update equations is used to estimate the process state by feedback control. A 2-D KF and a 3-D KF were designed to predict the location of new POS when change is detected, and then a decomposition-based differential evolution algorithm was used to obtain the optimal population.

Rong \textit{et al.} \cite{rong2018multidirectional} presented a multidirectional prediction strategy (MDP) to enhance the performance of evolution algorithms. A number of representative individuals are selected via adaptive clustering, and the population is then classified into several clusters according to the distances between individuals. Subsequently, MDP constructs time series models based on the historical information provided by the representative individuals, which is used to predict a number of evolutionary directions. However, only the trajectories in the previous two environments are considered in MDP.

To reduce computing cost, Li \textit{et al.} \cite{li2019predictive} proposed a predictive strategy based on special points (SPPS) including feed-forward center points, boundary points, close-to-center points, close-to-boundary points and knee points. The special point set that eliminates useless individuals can make the predicted population track POF more accurately. Knee points were also adopted in \cite{zou2017prediction} and \cite{wei2020prediction} to facilitate the tracking ability.
Wu \textit{et al.} \cite{wu2015directed} introduced a directed search strategy to predict a new population, in which the moving direction of POS is estimated by the position of centroid points. Min \textit{et al.} \cite{min2017multiproblem} proposed an adaptive knowledge reuse framework based on multiproblem surrogates, which accelerated the convergence of expensive multiobjective optimization.

Evolutionary transfer optimization (ETO) is an emerging paradigm in prediction \cite{tan2021evolutionary}, \cite{gupta2017insights}. Da \textit{et al.} \cite{da2018curbing} proposed an adaptive transfer framework to utilize the similarity of black-box optimization problems online. Bali \textit{et al.} \cite{bali2019multifactorial} dynamically adapted the extent of transfer between different tasks based on the optimal mixing of probabilistic model. Jiang \textit{et al.} \cite{jiang2017transfer} proposed a dynamic multiobjective optimization method based on transfer learning (Tr-DMOEA), which maps the POF in the past environment into a latent space via transfer component analysis (TCA), and then uses these mapped solutions to construct a high-quality population. Inspired by Tr-DMOEA, some transfer learning algorithms combined with memory mechanisms or pre-search strategies were proposed to solve DMOPs \cite{jiang2020individual,jiang2020knee,jiang2020fast}.

Differential models and linear models are often used in prediction algorithms. Liu \textit{et al.} \cite{liu2015integration} proposed an improved adaptive differential evolution crossover operator to facilitate population evolution, and use the information from the past two searches to make prediction. In \cite{zhou2007prediction}, Zhou \textit{et al.} constructed a time series for each individual in the population, and used a simple linear model to predict the individual position in the next time window. Cao \textit{et al.} \cite{cao2019decomposition} introduced a first-order and second-order mixed difference model based on the historical position to predict the centroid position of the population. Liang \textit{et al.} \cite{liang2019hybrid} proposed a hybrid of memory and prediction strategies for dynamic multiobjective optimization (MOEA/D-HMPS). In response to dissimilar changes, MOEA/D-HMPS exploits the moving direction of the population center at the previous two continuous time steps to predict the moving trajectory at the next moment. 

Prediction-based methods can take advantage of the trending in POS changes and show a promising performance in solving DMOPs. However, most prediction models assume that a linear correlation exists in the solutions at different times. While in many cases, even POS at adjacent moments are nonlinearly correlated. In addition, the time series constructed by most models is very short, and the information contained in the earlier searches is lost, which will affect the accuracy of prediction. Therefore, it is necessary to improve the generalization ability and information extraction ability of the prediction model. In this paper, incremental support vector machine (ISVM) and synthetic minority over-sampling technique (SMOTE) will be introduced to enhance the prediction model.

\section{Incremental Support Vector Machine and Synthetic Minority Over-sampling Technique}
\label{sec:ISVM AND SMOTE}

As incremental support vector machine and synthetic minority over-sampling technique are two critical components in the proposed algorithm, we will review them in this section for the completeness of the presentation. In the process of solving DMOPs, ISVM can incrementally learn knowledge from previous POS to accurately determine the quality of the solutions, while SMOTE can extract more information from a limited number of POS samples.

\subsection{Incremental Support Vector Machine}
Support Vector Machine (SVM) is a widely used binary classification model with sparsity and robustness \cite{zheng2020improved}. The strategy of SVM is to build an optimal hyperplane in the feature space for binary classification by maximizing the classification interval \cite{boser1992training}. The application of kernel function in SVM enables it to solve nonlinear and high dimensional pattern recognition problems well, which maps the samples from low dimensional space to a high dimensional space and turns the problem into a linearly separable one.

Generally, the training data of a typical SVM are imported in batch. But in some instances, SVM needs to be trained online. Incremental Support Vector Machine (ISVM) is proposed to handle incoming samples. ISVM can gradually update the parameters to accommodate new samples without training on all samples repeatedly \cite{laskov2006incremental}. Next, the principle of ISVM is briefly introduced.

To generate an ISVM, we need to create a discriminant function $f(x)=w \cdot \phi(x)+b$ learned from the samples $\left\{\left(x_{i}, y_{i}\right) \in \mathbb{R}^{m} \times\{-1,1\}, \forall i \in\{1, \ldots, N\}\right\}$. That is to solve a quadratic programming problem:
\begin{equation}
\left\{\begin{array}{ll}
\textbf{min} _{w, \mathrm{b}} & \frac{1}{2}\|w\|^{2}+C \cdot \sum_{i=1}^{N} \varepsilon_{i} \\
\textbf { s.t. } & y_{i}\left(w \cdot x_{i}+b\right) \geq 1-\varepsilon_{i}, i \in\{1, \ldots, N\}
\end{array}\right.
\end{equation}

The first term represents the maximized interval distance, while the second term is the regularization term. $C$ is the penalty parameter, and $\varepsilon_{i}$ is the slack variable used to build a soft margin. When dealing with nonlinear issues, the quadratic program is typically expressed in its dual form:
\begin{equation}
\min _{0 \leq \alpha_{i} \leq C}: L=\frac{1}{2} \sum_{i, j} \alpha_{i} Q_{i j} \alpha_{j}-\sum_{i} \alpha_{i}+b \sum_{i} y_{i} \alpha_{i}
\end{equation}
where ${Q_{ij}} = {y_i}{y_j}K({{{x}}_i},{{{x}}_j})$, $K({{{x}}_i},{{{x}}_j})= \varphi \left ( x_i \right )\cdot \varphi \left ( x_j \right ) $. $K$ is the kernel function that implicitly maps the vectors to a high dimensional feature space meanwhile simplify the calculation. The dual form of SVM discriminant function is herein represented as $f({x})=\sum_{j} \alpha_{j} y_{j} K\left({x}_{j}, {x}\right)+b$.

The Karush-Kuhn-Tucker (KKT) conditions uniquely define the solution of dual parameters $\left\{\alpha,b\right\}$ by the first-order conditions on $L$:
\begin{equation}
\begin{aligned}
G_{i} &=\frac{\partial L}{\partial \alpha_{i}}=\sum_{j} Q_{i j} \alpha_{j}+y_{i} b-1\left\{\begin{array}{ll}
>0,\alpha_{i}=0 \\
=0,0 \leq \alpha_{i} \leq C \\
<0,\alpha_{i}=C
\end{array}\right.\\
&\frac{\partial L}{\partial b}=\sum_{j} y_{j} \alpha_{j}=0
\end{aligned}
\end{equation}

The KKT conditions partition the training samples into three categories: the set $S$ of margin support vectors with $G_{i}=0$, the set $E$ of error support vectors with $G_{i}\le 0$ and the set $R$ of the remaining vectors with $G_{i} > 0$ \cite{cauwenberghs2000incremental}.

With the continuous introduction of new samples in the incremental learning process, the margin vector coefficients change simultaneously to keep the KKT conditions satisfied for all previously trained samples. For a new sample $m$ considered as a candidate support vector, the KKT conditions can be expressed differentially as:
\begin{equation}
\begin{array}{l}
\Delta G_{i}=Q_{i m} \Delta \alpha_{m}+\sum_{j \in S} Q_{i j} \Delta \alpha_{j}+y_{i} \Delta b, \forall i \in D \cup\{m\} \\
0=y_{m} \Delta \alpha_{m}+\sum_{j \in S} y_{j} \Delta \alpha_{j}
\end{array}
\end{equation}

Since $G_{i} = 0$ for the margin vector set $S = \left\{S_{1}, ..., S_{ls}\right\}$, the changes in coefficients must satisfy:
\begin{equation}
\mathcal{Q} \cdot\left[\begin{array}{c}
\Delta b \\
\Delta \alpha_{s_{1}} \\
\vdots \\
\Delta \alpha_{s_{l_{S}}}
\end{array}\right]=-\left[\begin{array}{c}
y_{m} \\
Q_{s_{1} m} \\
\vdots \\
Q_{s_{l_{S}} m}
\end{array}\right] \Delta \alpha_{m}
\end{equation}
where $\mathcal{Q}$ is a symmetric but not positive-definite Jacobian matrix:
\begin{equation}
\mathcal{Q}=\left[\begin{array}{cccc}
0 & y_{s_{1}} & \cdots & y_{s_{\ell_{S}}} \\
y_{s_{1}} & Q_{s_{1} s_{1}} & \cdots & Q_{s_{1} s_{\ell_{S}}} \\
\vdots & \vdots & \ddots & \vdots \\
y_{s_{\ell_{S}}} & Q_{s \ell_{S}} s_{1} & \cdots & Q_{s_{\ell_{S}} s_{\ell_{S}}}
\end{array}\right]
\end{equation}

Then we can get
\begin{equation}
\begin{aligned}
\Delta b &=\beta \Delta \alpha_{m} \\
\Delta \alpha_{j} &=\beta_{j} \Delta \alpha_{m}, \quad \forall j \in D
\end{aligned}
\end{equation}
with coefficient sensitivities
\begin{equation}
\left[\begin{array}{c}
\beta \\
\beta_{s_{1}} \\
\vdots \\
\beta_{s_{\ell_{S}}}
\end{array}\right]=-\mathcal{R} \cdot\left[\begin{array}{c}
y_{m} \\
Q_{s_{1} m} \\
\vdots \\
Q_{s_{\ell_{S}} m}
\end{array}\right]
\end{equation}
where $\mathcal{R}=\mathcal{Q}^{-1}$ and $\beta_{j}=0$ for all $j$ outside $S$. Hence, we have the KKT conditions in equation (7) changed according to:
\begin{equation}
\begin{array}{c}
\Delta G_{i}=\gamma_{i} \Delta \alpha_{m}, \forall i \in D \cup\{m\} \\
\gamma_{i}=Q_{i m}+\sum_{j \in S} Q_{i j} \beta_{j}+y_{i} \beta, \forall i \notin S
\end{array}
\end{equation}

To append a candidate vector $m$ into the margin vector set $S$, $\mathcal{R}$ is expanded as: 
\begin{equation}
\mathcal{R} \leftarrow\left[\begin{array}{cc}
 &0\\
\mathcal{R} & \vdots \\
 &0\\
0 \cdots 0 & 0
\end{array}\right]+\frac{1}{\gamma_{m}}\left[\begin{array}{c}
\beta \\
\beta_{s_{1}} \\
\vdots \\
\beta_{s_{\ell_{S}}} \\
1
\end{array}\right] \cdot\left[\beta, \beta_{s_{1}} \cdots \beta_{s_{\ell_{S}}}, 1\right]
\end{equation}

Conversely, to remove a vector $k$ from $S$, $\mathcal{R}$ is deflated as: 
\begin{equation}
\mathcal{R}_{i j} \leftarrow \mathcal{R}_{i j}-\mathcal{R}_{k k}^{-1} \mathcal{R}_{i k} \mathcal{R}_{k j}, \forall i, j \in S \cup\{0\} ; i, j \neq k
\end{equation}

Consequently, when a new sample $m$ is added to the training data set $D:D^{l+1}=D^{l}\cup\{m\}$, the solution of parameters $\left\{\alpha,b\right\}$ is updated with respect to the candidate $x_{m}, y_{m}$, the present solution and Jacobian inverse matrix $\mathcal{R}$. The incremental procedure can be summarized as:

1. Initialize $\alpha_{m}$ to zero, calculate $G_{m}$;

2. If $G_{m}> 0$, terminate ($m$ is not a margin or error vector);

3. If $G_{m}\le 0$, apply the largest possible increment $\alpha_{m}$ so that one of the following conditions occurs: (a) $G_{m}= 0$: Add $m$ to margin set $S$, update $\mathcal{R}$ accordingly and terminate; (b) $\alpha_{m}= C$: Add $c$ to error set $E$, terminate; (c) Elements of $D^{l}$ migrate across $S$, $E$ and $R$: update membership of elements and repeat step 3. If S changes, update $\mathcal{R}$ accordingly.

\subsection{Synthetic Minority Over-sampling Technique}
Uniformity in quantity of samples is a key factor affecting the accuracy of a classifier. Excessive differences in the number of samples may bias the classification results toward the dominant category. Synthetic minority over-sampling technique (SMOTE) is an effective way to deal with imbalanced data \cite{chawla2002smote}. 

\begin{figure}[h]
\centering
\includegraphics[width=\linewidth]{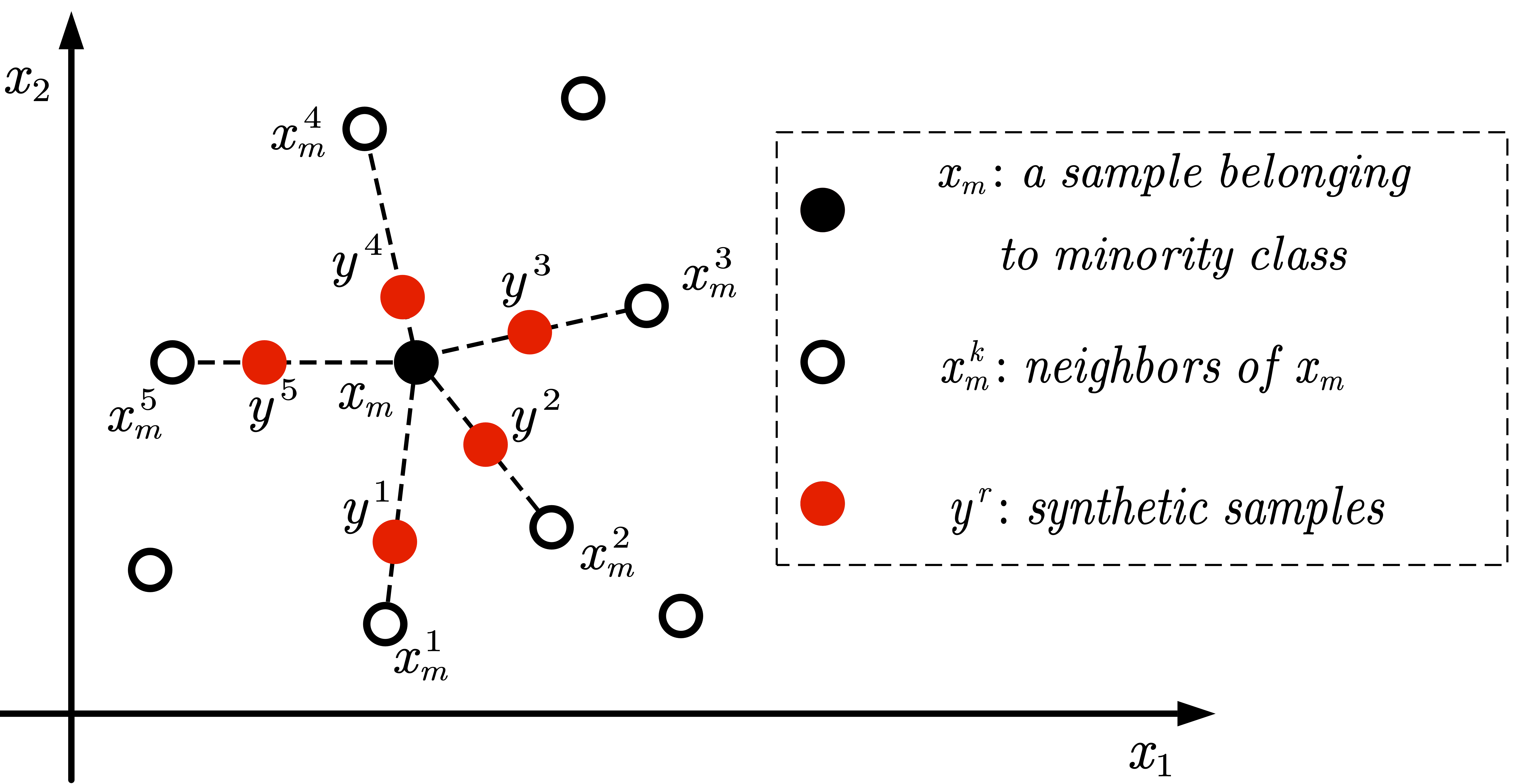}
\caption{Illustration of synthetic minority over-sampling technique (taking 2D decision space as an example).}
\label{fig:SMOTE}
\end{figure}

SMOTE suggests that there are some potentially available samples between the adjacent minority samples in the feature space. New samples are synthesized based on adjacent samples to balance the data set. As shown in Fig.~\ref{fig:SMOTE}, $x_{m}$ is a sample belonging to minority class in a two-dimensional feature space, and $X_m= \left\{x{_m^1},x{_m^2},…,x{_m^k}\right\}$ are $k$ neighbors of $x_{m}$ in the same category. By multiplying the difference of the two vectors by a random number (range 0-1), new samples $Y=\left\{y^{1},y^{2},…,y^{r}\right\}$ are synthesized along the line segment between $x_{m}$ and samples in $X_{m}$. The number of synthesized samples is determined by the oversampling rate $r$. Finally, $r$ new samples are added to the initial minority sample set to maintain the balance of sample class thereby improve the generalization ability of the classifier.

\section{Proposed Algorithm}
\label{sec:Proposed Algorithm}

\begin{figure*}[htbp]
\centering
\includegraphics[width=\textwidth]{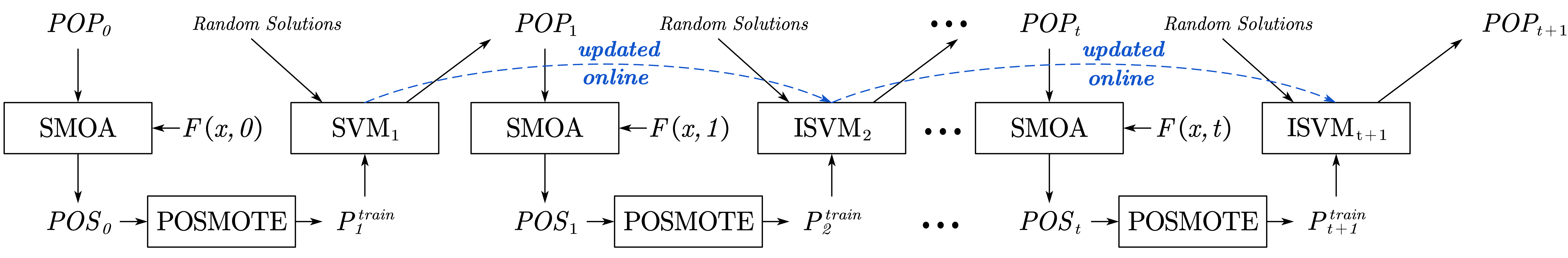}
\caption{Schematic of ISVM-DMOEA. At time $t$, the search for POS can be decomposed into: sampling (POSMOTE) $\to$ prediction (ISVMPRE) $\to$ optimization (SMOA).}
\label{fig:Schematic diagram}
\end{figure*}

In this section, we propose an online algorithm based on incremental support vector machine to solve DMOPs. The main idea is to train an ISVM classifier online by reusing the optimal solutions obtained in the previous environments  which are preprocessed by a sampling strategy based on SMOTE. When environmental changes are detected, the classifier predicts a high-quality initial population, which helps the optimization algorithm to converge to the true POS more quickly and accurately. The schematic diagram of the proposed ISVM-DMOEA is presented in Fig.~\ref{fig:Schematic diagram}. Briefly, ISVM-DMOEA consists of two main subroutines:1) POSMOTE sampling strategy and 2) ISVMPRE online prediction strategy. The description of subroutines is followed by the framework illustration of ISVM-DMOEA.

\subsection{POSMOTE Sampling Strategy}
As mentioned above, an ISVM classifier will be trained to predict an initial population for the next moment based on previous distribution of solutions. The optimal solutions obtained in the past environments are regarded as positive samples, while solutions with poor quality are regarded as negative samples accordingly. In DMOPs, a limited number of positive samples composed by optimal solutions will lead to low accuracy of ISVM. Therefore, it is necessary to generate a sufficient and balanced sample set for the predictor in advance. 

\begin{figure}
\centering
\includegraphics[width=\linewidth]{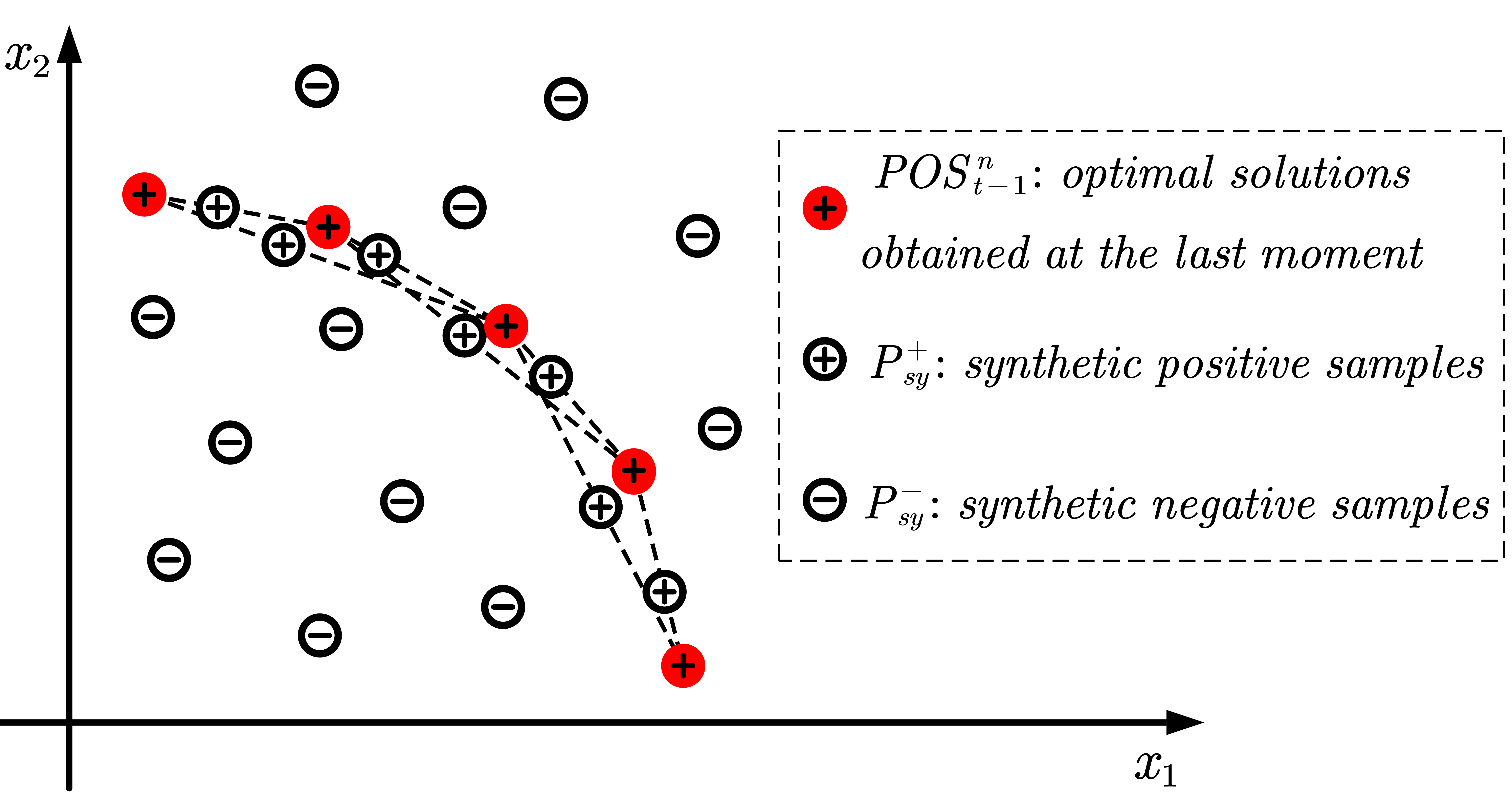}
\caption{Illustration of POSMOTE sampling strategy (taking 2D decision space as an example): positive samples are synthesized by calculating the linear interpolation between the solutions in $POS_{t-1}$, and negative samples are randomly generated according to the number of positive samples.}
\label{fig:POSMOTE}
\end{figure}

SMOTE is employed in the oversampling of positive samples belonging to minority class. The optimal solution set obtained at the last moment is expressed as $POS_{t-1}=\left\{POS{_{t-1}^1},POS{_{t-1}^2},…,POS{_{t-1}^n}\right\}$. In the first step, $k$ nearest neighbors of each optimal solution $POS{_{t-1}^n}$ are identified in $POS_{t-1}$ by the Euclidean distance. To synthesize a new sample $P{_{sy}^+}$, a vector $POS{_{t-1}^m}$ is selected from the $k$ nearest neighbors of $POS{_{t-1}^n}$. The attributes of $P{_{sy}^+}$ in each dimension are calculated randomly by linear interpolation of $POS{_{t-1}^m}$ and $POS{_{t-1}^n}$:
\begin{equation}
\begin{split}
P_{s y}^{+}(d)={POS}_{t-1}^{n}(d)+{Rand} \times \left({POS}_{t-1}^{n}(d)-{POS}_{t-1}^{m}(d)\right)
\end{split}
\end{equation}
where $d \in\{1, \ldots, D\}$, and $D$ is the dimension of the decision vector. $Rand$ represents a random number in (0,1). The number of new samples synthesized by each optimal solution depends on the oversampling rate $r$. $n\times r$ synthetic samples and $n$ original samples constitute the positive sample set $P^{+}$.

Compared with positive samples, the number of negative samples is very large. The generation of negative sample set can be regarded as a down-sampling process. Since most of the solutions in the decision space fall outside of the $POS_{t-1}$, negative sample set can be composed of randomly synthesized solutions. To generate a balanced sample set,  $n\times (r+1)$ negative samples are synthesized. As illustrated in Fig.~\ref{fig:POSMOTE}, the number of negative samples in $P^{-}$ is consistent with that of positive samples in $P^{+}$, and the samples are sufficient for the training of ISVM. The details of POSMOTE are shown in Algorithm ~\ref{alg:POSMOTE}.

\begin{algorithm}
	\caption{POSMOTE --- Sampling Strategy}
     \label{alg:POSMOTE}
     \KwIn{ the POS with $n$ solutions obtained at the last moment, $POS_{t-1}$; the oversampling rate $r$; the number of nearest neighbors considered, $k$; the number of decision variables, $D$;}
     \KwOut{a balanced sample set, $P_{train}$;}
   initialize the sample set, $P_{train}$ = $\emptyset$\;
    \For{i = \rm 1 to $n$}{
         identify  $k$  nearest neighbors of $POS_{t-1}^i$\;
     \While{$r\ne0$}{
         randomly select a neighbor in $k$ neighbors\;     
         \For{j = \rm 1 to $D$}{   
              calculate the attributes of the synthesized vector $P_{sy}^+$ in each dimension according to formula 16\;
         }
          $P^+$ = $P^+ \cup P_{sy}^+$\;
         $r$ = $r$ - 1\;
}
     }
     $P^+$= $P^+ \cup POS_{t-1}$\;
     \For{i = \rm 1 to $n(r+1)$}{
      randomly generate a vector $P_{sy}^-$ in the decision space\;
     $P^-$= $P^- \cup P_{sy}^-$\;
}
$P_{train}$ = \{$P^+$, $P^-$\}\;
\Return $P_{train}$\;
\end{algorithm}

\subsection{ISVMPRE Online Prediction Strategy}
For a specific problem, we assume that a general distribution pattern lies in the optimal solutions at different instants. The strategy of prediction is to explore the general pattern with a binary classifier. A well-trained classifier can estimate whether a solution has the characteristics of being an optimal solution.

SVM can map the solutions in the decision space to a high dimensional feature space to construct a linear classifier, which can explore the implicit connection between the optimal solutions in the original space, regardless linear or nonlinear. However, if only optimal solutions in the former environment are used to construct an SVM each time, the information contained in solutions at earlier instants cannot be extracted; while if all the optimal solutions in previous environments are used to train the SVM, enormous amount of time and storage space will be needed.

To construct a prediction model with both high efficiency and accuracy, incremental SVM is employed. The parameters of ISVM are constantly updated as the environment changes. \rm ISVM$_{t}$  is an updated model of \rm ISVM$_{t-1}$ based on new samples imported online. A randomly generated solution is recognized as a candidate for optimal solution if it is classified as positive in \rm ISVM$_{t}$. To generate a promising population, \rm ISVM$_{t}$ works as a filter: random solutions classified as positive are retained, and the negative ones are discarded. Finally, $N_{P}$ positive samples are placed into $POP_{t}$, which is a high-quality initial population for the new environment. The pseudocode of ISVMPRE is presented in Algorithm ~\ref{alg:ISVMpredicton}.

\begin{algorithm}
	\caption{ISVMPRE --- Online Prediction Strategy}
     \label{alg:ISVMpredicton}
     \KwIn{ the population size, $N_{p}$; the parameters of ISVM at the last moment, \rm ISVM$_{t-1}$; the training samples, $P_{train}$;}
     \KwOut{a predicted population, $POP_{t}$; updated parameters of ISVM at time t, \rm ISVM$_{t}$; }
   initialize the predicted population: $POP_{t}$ = $\emptyset$\ ,$N_{count}$ = 0\; 
   incrementally train the \rm ISVM$_{t}$ based on \rm ISVM$_{t-1}$ with samples in $P_{train}$\;
 \While {$N_{count}$ \textless $N_{p}$}{
    randomly generate a vector $P_{rand}$ in the decision space\;
    get the label of $P_{rand}$ in \rm ISVM$_{t}$\;
    \If {\rm lable = +1}{
     $POP_{t}$ = $POP_{t} \cup P_{rand}$\;
    $N_{count}$ = $N_{count}$ + 1\;
}
}

\Return $POP_{t}$, \rm ISVM$_{t}$\;
\end{algorithm}

\subsection{Framework of ISVM-DMOEA}

Algorithm ~\ref{alg:Framwork of IISP} depicts the overall framework of the proposed ISVM-DMOEA. The entire process of solving a dynamic multiobjective optimization problem is accompanied by the online learning process of ISVM-based predictor. 

In the first environment, the population is initialized randomly, and then the initial population is optimized by a population-based static multiobjective optimization algorithm (SMOA) to obtain the corresponding $POS_0$. When changes in the environment are detected, a new sample set is synthesized based on $POS_{t-1}$ obtained in the previous environments. The number and characteristics of new samples are predefined by the parameters in POSMOTE. Subsequently, new samples are introduced into the prediction strategy, where the parameters of ISVM are updated online to incorporate more general information from previous POS. A high-quality initial population is then predicted by ISVMPRE and further optimized by SMOA until POS is obtained.

It is worth noting that, the procedures of POSMOTE sampling strategy, ISVMPRE prediction strategy and SMOA optimization are successive and non-interfering in ISVM-DMOEA, so any population-based multiobjective optimization algorithm can be embedded in the proposed framework.

\begin{algorithm}
	\caption{Framwork of ISVM-DMOEA}
     \label{alg:Framwork of IISP}
     \KwIn{ the dynamic optimization problem, $F(x,t)$; the population size, $N_{p}$; the oversampling rate $r$; the number of nearest neighbors considered, $k$; the number of decision variables, $D$;}
     \KwOut{The solutions at time t, $POS_{t}$;}
     randomly initialize a population $POP_{0}$, $t$ = 0\;
     $POS_{0}$ = SMOA($POP_{0},F(x,0),N_{p}$)\;
     \While {change detected}{
     $t$ = $t$ + 1\;
     $P_{train}$ = POSMOTE($POS_{t-1}$, $r$, $k$, $D$)\;
     $POP_{t}$ = ISVMPRE($P_{train}$, \rm ISVM$_{t-1}$, $N_{p}$ )\;
     $POS_{t}$ = SMOA($POP_{t}$, $F(x,t)$, $N_{p}$)\;
}

\Return $POS_{t}$\;
\end{algorithm}

\section{Experimental Study}
\label{sec:Experimental Study}

\subsection{Benchmark Problems and performance indicators}

In this paper, the performance of the proposed ISVM-DMOEA and chosen competing algorithms is uniformly examined on CEC 2018 DMO benchmark suite with nine bi-objective and five tri-objective problems. These problems are named DF1-DF14, which cover diverse problem characteristics such as dynamic POF/POS geometries, irregular POF shapes, variable linkage and disconnectivity. The time instance t involved in the problems is defined as $t=\frac{1}{n_{t}}\left\lfloor\tau / \tau_{t}\right\rfloor$, where $n_{t}$, $\tau_{t}$ and $\tau$ denote the severity of change, the frequency of change and the iteration counter, respectively. For each problem, various pairs of $n_{t}$ and $\tau_{t}$ are implemented to configure different dynamic characteristics. The definition of test instances is detailed in \cite{jiang2018benchmark}.

The following matrix are adopted to assess the performance of algorithms:

1) Inverted Generational Distance (IGD): IGD is a frequently-used metric to measure the convergence and diversity of the solutions by computing the difference between true POF and the POF estimated by an algorithm. At time $t$, IGD is calculated as:
\begin{equation}
\mathrm{IGD}\left(\mathrm{POF}_{t}^{*}, \mathrm{POF}_{t}^{e}\right)=\frac{\sum_{p \in \mathrm{POF}_{t}^{*}} d\left(p, \mathrm{POF}_{t}^{e}\right)}{\left|\mathrm{POF}_{t}^{*}\right|}
\end{equation}
where POF$_{t}^{*}$ is a set of points uniformly sampled from the true POF$_{t}$, and POF$_{t}^{e}$ is estimated by the tested algorithm. $d(p,\mathrm{POF}_{t}^{e})$ represents the minimum Euclidian distance between $p$ belonging to POF$_{t}^{*}$ and the points in POF$_{t}^{e}$. The $d(p,\mathrm{POF}_{t}^{e})$ of each point in POF$_{t}^{*}$ is calculated and then the average distance is computed.

To apply IGD to the changing environments, a variant of IGD called MIGD is adopted. MIGD takes the average of the IGD values at different time, given by:
\begin{equation}
\mathrm{MIGD}=\frac{\sum_{t \in T} \mathrm{IGD}\left(\mathrm{POF}_{t}^{*}, \mathrm{POF}_{t}^{e}\right)}{|T|}
\end{equation}
where T is a set of discrete instants and $\left|T\right|$ is the number of changes in a run. In this paper, 1000 points are uniformly sampled from POF$_{t}^{*}$ for the calculation of IGD.

2) Hypervolume (HV): HV is used to evaluate the diversity and distribution of the solutions by computing the hypervolume of the region dominated by the obtained POF\cite{tian2017indicator}. HV is formally defined as follows:
\begin{equation}
\mathrm{HV}(\mathrm{POF}_{t}^{e}, rp)=\Lambda\left(\bigcup_{p \in \mathrm{POF}_{t}^{e}}\left\{q \mid rp \succ q \succ p\right\}\right)
\end{equation}
where $\Lambda$ denotes the Lebesgue measure and $rp \in R^D$ is the reference point calculated by the maximum value of the $d-th$ objective of the POF. Similarily, we can define a metric MHV based on HV, which is given as: 
\begin{equation}
\mathrm{MHV}=\frac{\sum_{t \in T} \mathrm{HV}\left(\mathrm{POF}_{t}^{e}, rp\right)}{|T|}
\end{equation}

\subsection{Compared algorithms and parameter settings}

Without loss of generality, we choose the regularity model-based multiobjective estimation of distribution algorithm (RM-MEDA) as the static optimizer in the proposed ISVM-DMOEA. RM-MEDA is a widely-used algorithm that constructs a posterior probability distribution model of promising solutions based on global statistical information extracted from selected solutions \cite{zhang2008rm}. The proposed algorithm incorporating RM-MEDA is called ISVM-RM-MEDA.

To verify the performance of ISVM-RM-MEDA, four state-of-the-art algorithms are chosen for comparison: Kalman prediction-based MOEA (MOEA/D-KF) \cite{muruganantham2015evolutionary}, population prediction strategy (PPS)\cite{zhou2013population}, support vector regression-based predictor (MOEA/D-SVR)\cite{cao2019evolutionary} and transfer learning-based DMOEA (Tr-DMOEA)\cite{jiang2017transfer}. These algorithms use different prediction strategies to solve DMOPs, and have achieved considerable performance. Besides, RM-MEDA is modified to adapt to dynamic problems, called DA-RM-MEDA. For a fair comparison, the baseline optimizer in these algorithms are uniformly replaced by RM-MEDA. In our empirical studies, the compared algorithms are referred to as KF-RM-MEDA, PPS-RM-MEDA, SVR-RM-MEDA, TR-RM-MEDA and DA-RM-MEDA, respectively.

The parameters in the test environments and algorithms are set as follows:

1) Population size and number of decision variables: The population size is set to be 100 for bi-objective problems (DF1-DF9) and 150 for tri-objective problems (DF10-DF14). The number of decision variables is set as 10 for all test problems. 

2) Dynamic characteristics: For each problem, three pairs of dynamic characteristics are set by different combinations of $n_{t}$ and $\tau_{t}$: ( $n_{t}$=10, $\tau_{t}$=10); ( $n_{t}$=5, $\tau_{t}$=10); ( $n_{t}$=10, $\tau_{t}$=5). $\tau / \tau_{t}$ is fixed to be 30, which ensures there are 30 changes in each run. Each algorithm is run 20 times on each test instance independently.

3) Parameters in algorithms: In the proposed algorithm, Gaussian RBF kernel is selected as the kernel function of ISVM, and the kernel scale is determined by grid search; both the oversampling rate $r$ and the number of $k$ nearest neighbors are both set to be 5 in POSMOTE. The parameters of RM-MEDA and compared algorithms are referenced from the original papers.

\subsection{Comparison Study}

The statistical results of MIGD and MHV obtained by ISVM-RM-MEDA and five competing algorithms are presented in Table ~\ref{table:MIGD comparision} and Table ~\ref{table:MHV comparision}, respectively. For an intuitive comparison, the best values on each instance are hilighted in bold and the Wilcoxon rank sum test at the 0.05 significance level is carried out to indicate the significance between different results.

It can be seen from Table~\ref{table:MIGD comparision} that ISVM-RM-MEDA obtains the best MIGD results on the majority of the tested problems, implying that the proposed algorithm has superior tracking ability of dynamic POF in most cases. ISVM-RM-MEDA performs slightly worse than PPS-RM-MEDA in half of the configurations on DF1, DF2 and DF9, which may be attributed to the comprehensive strategy of PPS combining POS center prediction and manifold estimation. The POS of DF8 contains a group of stationary centroids and the POS varies nonlinearly over time, which makes it very difficult to be approximated. Therefore, the MIGD values on DF8 obtained by the randomly reinitialized approach for RM-MEDA (DA-RM-MEDA) is better than those of all the algorithms based on prediction. Nevertheless, two nonlinear prediction driven algorithms, ISVM-RM-MEDA and SVR-RM-MEDA, achieve better results than other prediction algorithms on DF8. ISVM-RM-MEDA seems slightly inefficient on the complex problem DF12 with a time-varying number of POF holes. Except for DF12, ISVM-RM-MEDA significantly outperform other algorithms on tri-objective problems. Obviously, the MIGD values obtained by ISVM-RM-MEDA are competitive in three pairs of ($n_{t}$ ,$\tau_{t}$) configurations, demonstrating that ISVM-RM-MEDA is a robust model with less sensitive to the frequency and severity of change.

\begin{table*}[htbp]
  \centering
  \caption{MEAN AND STANDARD DEVIATION VALUES OF MIGD OBTAINED BY ISVM-RM-MEDA AND COMPARED ALGORITHMS}
\label{table:MIGD comparision}
\resizebox{\textwidth}{8cm}{
\begin{threeparttable} 
    \begin{tabular}{cccccccc}
    \toprule
    Prob & ($n_{t}$ ,$\tau_{t}$) & DA-RM-MEDA & KF-RM-MEDA & PPS-RM-MEDA & SVR-RM-MEDA & TR-RM-MEDA & ISVM-RM-MEDA \\
    \midrule
    \multirow{3}[2]{*}{DF1} & (10,10) & 0.1576±4.55E-03(+) & 0.1972±1.99E-02(+) & \textbf{0.0365±7.34E-03(-)} & 0.1305±9.37E-03(+) & 0.0929±1.48E-02(+) & 0.0424±2.02E-03 \\
          & (5,10) & 0.1579±7.64E-03(+) & 0.1958±2.32E-02(+) & 0.0717±6.06E-03(+) & 0.1544±1.38E-02(+) & 0.0876±1.14E-02(+) & \textbf{0.0674±1.06E-02} \\
          & (10,5) & 0.3481±9.82E-03(+) & 0.2462±7.21E-03(+) & \textbf{0.0949±2.11E-02(-)} & 0.2683±1.45E-02(+) & 0.1672±7.29E-03(=) & 0.1516±6.24E-02 \\
    \midrule
    \multirow{3}[2]{*}{DF2} & (10,10) & 0.0982±2.56E-03(+) & 0.1437±1.83E-02(+) & 0.0408±8.58E-03(+) & 0.0854±3.44E-03(+) & 0.0658±1.05E-02(+) & \textbf{0.0386±1.53E-03} \\
          & (5,10) & 0.1034±1.41E-03(+) & 0.1366±1.29E-02(+) & 0.0741±5.64E-03(+) & 0.0933±7.11E-03(+) & 0.0711±7.32E-03(+) & \textbf{0.0634±3.92E-03} \\
          & (10,5) & 0.2286±4.12E-03(+) & 0.2018±8.95E-03(+) & \textbf{0.0917±7.86E-03(-)} & 0.1742±8.74E-03(+) & 0.1262±1.40E-02(+) & 0.1018±5.53E-03 \\
    \midrule
    \multirow{3}[2]{*}{DF3} & (10,10) & 0.4155±9.69E-03(+) & 0.2686±1.55E-02(+) & 0.2012±6.02E-03(=) & 0.4203±1.49E-02(+) & 0.2534±1.15E-02(+) & \textbf{0.1991±8.20E-03} \\
          & (5,10) & 0.4238±1.15E-02(+) & 0.2656±1.84E-02(+) & 0.2415±1.54E-02(+) & 0.4089±1.63E-02(+) & 0.2360±1.19E-02(+) & \textbf{0.2251±1.57E-02} \\
          & (10,5) & 0.5371±1.74E-02(+) & 0.3276±8.79E-03(+) & 0.2809±3.57E-02(+) & 0.5361±2.29E-02(+) & 0.3810±3.96E-02(+) & \textbf{0.2687±3.38E-02} \\
    \midrule
    \multirow{3}[2]{*}{DF4} & (10,10) & 1.4078±9.79E-03(+) & 2.0779±2.56E-02(+) & 1.0118±1.78E-02(+) & 1.3655±3.07E-02(+) & 1.2846±3.51E-02(+) & \textbf{0.9658±3.47E-03} \\
          & (5,10) & 1.4163±1.05E-02(+) & 2.0865±1.64E-02(+) & 1.0334±1.15E-02(+) & 1.3735±2.32E-02(+) & 1.3092±3.96E-02(+) & \textbf{0.9783±7.70E-03} \\
          & (10,5) & 1.7953±2.38E-02(+) & 2.1402±1.57E-02(+) & 1.0861±1.48E-02(=) & 1.7378±4.49E-02(+) & 1.5536±6.89E-02(+) & \textbf{1.0365±8.62E-03} \\
    \midrule
    \multirow{3}[2]{*}{DF5} & (10,10) & 1.5713±2.82E-02(+) & 5.0891±1.03E+00(+) & 1.1942±1.30E-02(+) & 1.8507±3.50E-02(+) & 2.1915±1.33E-01(+) & \textbf{1.0040±1.12E-02} \\
          & (5,10) & 1.7356±2.23E-02(+) & 5.4275±7.56E-01(+) & 1.1311±6.80E-03(=) & 1.7124±2.94E-02(+) & 1.8596±1.60E-01(+) & \textbf{1.1293±8.90E-03} \\
          & (10,5) & 1.9029±1.84E-02(+) & 4.4369±1.05E+00(+) & 1.3787±6.25E-02(+) & 2.1661±3.93E-02(+) & 2.1836±3.44E-02(+) & \textbf{1.1440±1.74E-02} \\
    \midrule
    \multirow{3}[2]{*}{DF6} & (10,10) & 7.5990±5.43E-01(+) & 26.517±2.48E+00(+) & \textbf{3.0892±3.07E-01(-)} & 6.1963±2.24E-01(+) & 4.8816±2.87E-01(+) & 3.5468±1.99E-01 \\
          & (5,10) & 7.4662±3.42E-01(+) & 28.525±3.65E+00(+) & 6.2691±6.85E-01(+) & 6.5605±3.20E-01(+) & \textbf{2.9575±2.29E-01(-)} & 5.1673±4.55E-01 \\
          & (10,5) & 10.794±4.46E-01(+) & 27.334±1.47E+00(+) & 6.5176±5.38E-01(+) & 8.7349±5.45E-01(+) & 6.7721±3.09E-01(+) & \textbf{6.4605±8.49E-01} \\
    \midrule
    \multirow{3}[2]{*}{DF7} & (10,10) & 6.2632±3.93E-01(+) & 26.876±2.69E-01(+) & \textbf{2.9774±2.83E-01(-)} & 5.0264±4.13E-01(+) & 4.2575±1.90E-01(+) & 3.9424±1.81E-01 \\
          & (5,10) & 6.5309±3.50E-01(+) & 24.683±3.55E+00(+) & 5.0051±4.46E-01(+) & 5.6358±4.20E-01(+) & \textbf{2.3972±1.66E-01(-)} & 4.4235±7.12E-01 \\
          & (10,5) & 9.3613±5.68E-01(+) & 24.894±3.67E+00(+) & 5.7018±5.14E-01(+) & 7.5320±3.68E-01(+) & 5.9953±5.53E-01(+) & \textbf{5.6874±7.62E-01} \\
    \midrule
    \multirow{3}[2]{*}{DF8} & (10,10) & \textbf{0.7128±1.42E-02(-)} & 1.3389±1.00E-02(+) & 0.8667±7.93E-03(+) & 0.7401±1.06E-02(-) & 0.8182±1.17E-02(+) & 0.7672±1.23E-02 \\
          & (5,10) & \textbf{0.7325±8.61E-03(-)} & 1.3879±4.88E-03(+) & 0.8947±1.14E-02(+) & 0.7768±1.07E-02(-) & 0.8283±1.32E-02(+) & 0.7957±2.33E-03 \\
          & (10,5) & \textbf{0.6897±6.85E-03(-)} & 1.3553±1.01E-02(+) & 0.7860±2.10E-02(+) & 0.7196±1.11E-02(-) & 0.8125±6.75E-03(+) & 0.7451±1.38E-02 \\
    \midrule
    \multirow{3}[2]{*}{DF9} & (10,10) & 2.3578±3.93E-02(+) & 3.3843±1.57E-02(+) & \textbf{1.5801±1.13E-02(-)} & 1.9235±5.45E-02(-) & 1.8126±4.12E-02(-) & 1.9743±2.32E-02 \\
          & (5,10) & 2.2928±3.12E-02(+) & 3.3336±7.23E-01(+) & \textbf{1.2775±4.44E-02(=)} & 1.6277±3.81E-02(+) & 1.3831±5.47E-02(+) & 1.2819±3.39E-02 \\
          & (10,5) & 2.4821±4.62E-02(+) & 3.3780±9.79E-03(+) & 1.6691±3.88E-02(+) & 2.0972±3.65E-02(+) & 1.8970±8.15E-02(+) & \textbf{1.5579±5.12E-02} \\
    \midrule
    \multirow{3}[2]{*}{DF10} & (10,10) & 0.1683±1.53E-03(+) & 0.2286±2.04E-03(+) & 0.1389±1.84E-03(+) & 0.1792±1.80E-03(+) & 0.1198±2.91E-03(+) & \textbf{0.1046±4.95E-04} \\
          & (5,10) & 0.1462±1.55E-03(+) & 0.2285±3.85E-03(+) & 0.1478±1.88E-03(+) & 0.1815±2.81E-03(+) & 0.1320±2.24E-03(+) & \textbf{0.0933±2.45E-03} \\
          & (10,5) & 0.2053±2.83E-03(+) & 0.3072±5.13E-03(+) & 0.1874±2.68E-03(+) & 0.2181±3.09E-03(+) & 0.1548±3.98E-03(+) & \textbf{0.1396±1.48E-03} \\
    \midrule
    \multirow{3}[2]{*}{DF11} & (10,10) & 0.1935±7.33E-03(+) & 0.2373±7.34E-03(+) & 0.1565±1.09E-02(+) & 0.2280±2.55E-03(+) & 0.3587±4.45E-02(+) & \textbf{0.0778±1.07E-03} \\
          & (5,10) & 0.1919±3.35E-03(+) & 0.2823±5.46E-03(+) & 0.2235±4.30E-03(+) & 0.2436±3.49E-03(+) & 0.4266±2.40E-02(+) & \textbf{0.0791±1.39E-03} \\
          & (10,5) & 0.3418±6.32E-03(+) & 0.3808±3.66E-03(+) & 0.1783±7.75E-03(+) & 0.3656±4.36E-03(+) & 0.3395±1.87E-02(+) & \textbf{0.1112±3.70E-03} \\
    \midrule
    \multirow{3}[2]{*}{DF12} & (10,10) & 0.7447±2.51E-02(-) & \textbf{0.4199±3.98E-03(-)} & 1.1756±5.61E-03(=) & 0.7261±2.11E-02(-) & 1.1853±0.00E+00(+) & 1.1654±4.96E-03 \\
          & (5,10) & 0.6314±1.29E-02(-) & \textbf{0.4229±7.75E-04(-)} & 1.1804±3.31E-03(=) & 0.6241±5.64E-03(-) & 1.1858±0.00E+00(+) & 1.1702±3.09E-03 \\
          & (10,5) & 0.6714±1.39E-02(-) & \textbf{0.4108±3.10E-03(-)} & 1.1760±5.61E-03(+) & 0.6718±7.88E-03(-) & 1.1853±1.18E-08(+) & 1.1428±7.09E-03 \\
    \midrule
    \multirow{3}[2]{*}{DF13} & (10,10) & 1.6407±2.98E-02(+) & 1.3463±2.28E-02(+) & 1.3815±2.58E-02(+) & 1.9113±4.83E-02(+) & 2.0254±4.36E-02(+) & \textbf{1.1329±2.78E-02} \\
          & (5,10) & 1.7724±3.20E-02(+) & 1.2424±1.84E-02(+) & 1.2986±2.03E-02(+) & 1.7943±2.12E-02(+) & 1.8959±2.20E-02(+) & \textbf{1.1945±3.35E-02} \\
          & (10,5) & 1.8983±2.36E-02(+) & 1.6241±4.30E-02(+) & 1.6325±3.88E-02(+) & 2.2068±4.07E-02(+) & 2.3661±3.62E-02(+) & \textbf{1.3567±4.16E-02} \\
    \midrule
    \multirow{3}[2]{*}{DF14} & (10,10) & 1.1099±3.00E-02(+) & 0.9028±1.43E-02(+) & 0.8579±7.83E-03(+) & 1.2956±3.70E-03(+) & 1.2527±3.54E-02(+) & \textbf{0.6884±6.92E-03} \\
          & (5,10) & 1.2009±1.91E-02(+) & 0.8273±1.53E-02(+) & 0.8362±6.64E-03(+) & 1.1804±1.96E-02(+) & 1.1503±1.86E-02(+) & \textbf{0.7745±9.52E-03} \\
          & (10,5) & 1.3038±2.59E-02(+) & 1.0438±1.96E-02(+) & 1.0334±4.22E-02(+) & 1.4983±5.73E-02(+) & 1.5649±3.39E-02(+) & \textbf{0.7589±1.20E-02} \\
    \bottomrule
    \end{tabular}
 \begin{tablenotes}
        \footnotesize
        \item (+), (=) and (-) indicate that ISVM-RM-MEDA performs significantly better or equivalently or worse than the compared algorithms, respectively.
      \end{tablenotes}
  \end{threeparttable}}
\end{table*}

\begin{figure*}[htbp]
\centering
\includegraphics[width=\textwidth]{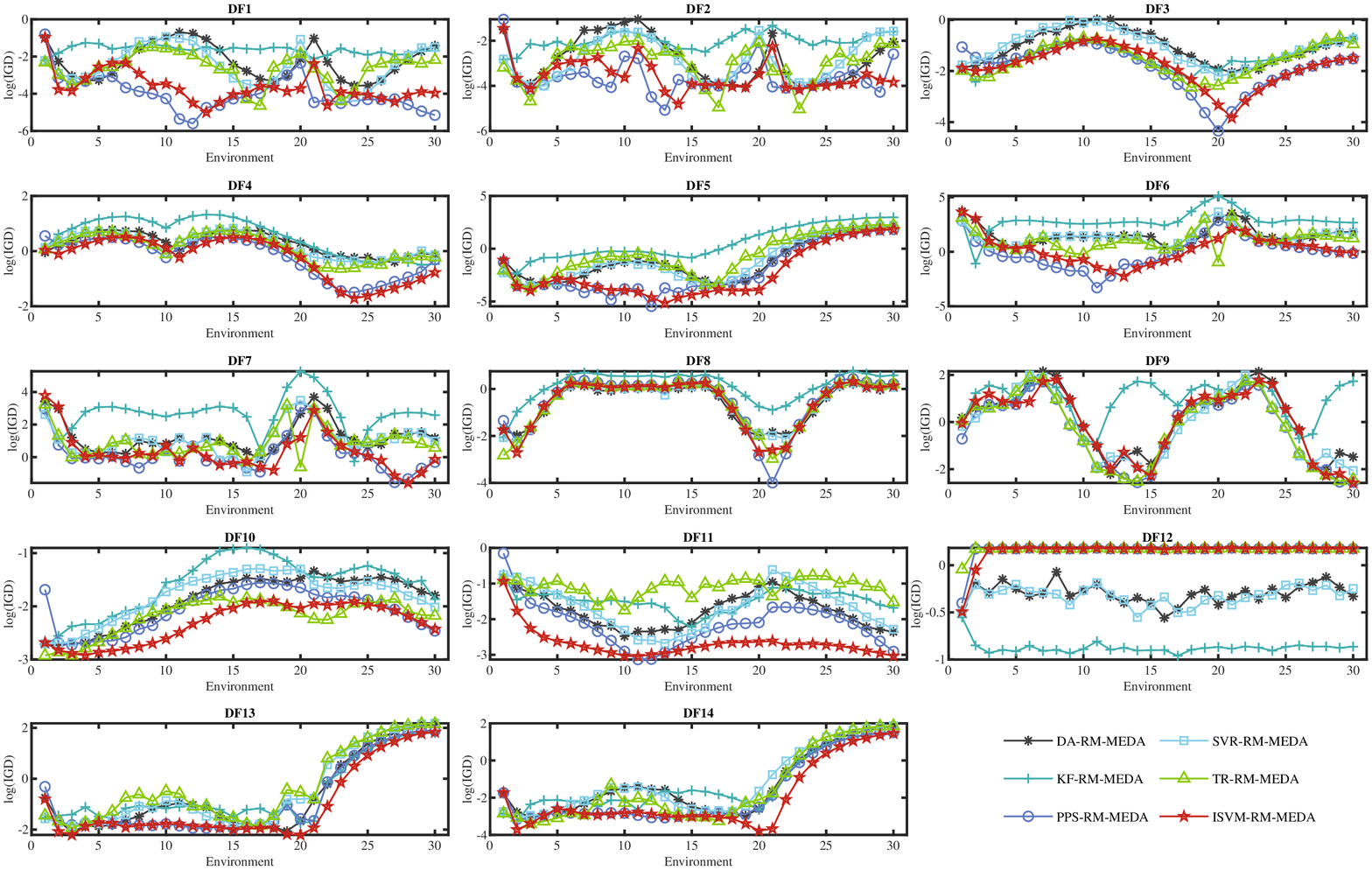}
\caption{Average IGD values versus environmental changes for different problems with $n_{t}$ = 10 and $\tau_{t}$ = 10.} 
\label{fig:EVOLUTIONIGD}
\end{figure*}

Fig.~\ref{fig:EVOLUTIONIGD} depicts the evolution curves of the IGD values on DF1-DF14 averaged over 20 runs with $n_{t}$ = 10 and $\tau_{t}$ = 10. We can see that the IGD curves obtained by ISVM-RM-MEDA are relatively low on most problems except for DF12. Besides, the fluctuation range of the curves obtained by ISVM-RM-MEDA is slighter than other algorithms, indicating that the proposed method can respond to the environmental changes quickly and stably.

As shown in Table~\ref{table:MHV comparision}, ISVM-RM-MEDA achieves the best performance on 30 out of 42 test instances in terms of MHV metrics. The MHV value obtained by ISVM-RM-MEDA is slightly worse than that of PPS-RM-MEDA on DF1-DF2 and TR-RM-MEDA on DF12, but better than that of the rest algorithms. The Wilcoxon test result indicates that there is no significant difference between these solution sets.

\begin{table*}[htbp]
  \centering
  \caption{MEAN AND STANDARD DEVIATION VALUES OF MHV OBTAINED BY ISVM-RM-MEDA AND COMPARED ALGORITHMS}
\label{table:MHV comparision}
\resizebox{\textwidth}{8cm}{
\begin{threeparttable} 
    \begin{tabular}{cccccccc}
    \toprule
    Prob & ($n_{t}$ ,$\tau_{t}$) & DA-RM-MEDA & KF-RM-MEDA & PPS-RM-MEDA & SVR-RM-MEDA & TR-RM-MEDA & ISVM-RM-MEDA \\
    \midrule
    \multirow{3}[2]{*}{DF1} & (10,10) & 0.4047±3.10E-03(+) & 0.3979±1.66E-02(+) & \textbf{0.5654±9.47E-03(=)} & 0.4311±8.91E-03(+) & 0.4868±1.57E-02(+) & 0.5593±3.22E-03 \\
          & (5,10) & 0.4258±4.95E-03(+) & 0.4146±2.29E-02(+) & \textbf{0.5352±7.54E-03(=)} & 0.4330±1.49E-02(+) & 0.5225±1.40E-02(=) & 0.5286±1.06E-02 \\
          & (10,5) & 0.2971±4.45E-03(+) & 0.3651±5.38E-03(+) & \textbf{0.4905±2.67E-02(-)} & 0.3175±9.91E-03(+) & 0.3989±5.19E-03(+) & 0.4514±5.43E-02 \\
    \midrule
    \multirow{3}[2]{*}{DF2} & (10,10) & 0.7138±3.28E-03(+) & 0.6364±2.14E-02(+) & \textbf{0.8058±1.04E-02(=)} & 0.7282±5.73E-03(+) & 0.7334±1.85E-02(+) & 0.7979±3.28E-03 \\
          & (5,10) & 0.7025±2.24E-03(+) & 0.6463±1.67E-02(+) & \textbf{0.7686±1.02E-02(=)} & 0.7167±9.97E-03(+) & 0.7305±1.57E-02(+) & 0.7578±4.66E-03 \\
          & (10,5) & 0.5820±4.43E-03(+) & 0.5598±7.89E-03(+) & \textbf{0.7119±1.28E-02(+)} & 0.5979±1.21E-02(+) & 0.6557±1.93E-02(+) & 0.6910±7.04E-03 \\
    \midrule
    \multirow{3}[2]{*}{DF3} & (10,10) & 0.2125±3.83E-03(+) & 0.2867±1.18E-02(+) & 0.3622±3.57E-03(+) & 0.2080±7.47E-03(+) & 0.2958±7.23E-03(+) & \textbf{0.3811±6.67E-03} \\
          & (5,10) & 0.2042±2.68E-03(+) & 0.2946±1.61E-02(+) & 0.3362±1.25E-02(+) & 0.2142±7.86E-03(+) & 0.3092±1.09E-02(+) & \textbf{0.3551±1.07E-02} \\
          & (10,5) & 0.1504±7.65E-03(+) & 0.2383±8.31E-03(+) & 0.2884±2.39E-02(=) & 0.1482±9.61E-03(+) & 0.1982±1.83E-02(+) & \textbf{0.3049±2.25E-02} \\
    \midrule
    \multirow{3}[2]{*}{DF4} & (10,10) & 1.8391±3.73E-02(+) & 1.2709±4.45E-02(+) & 3.3939±5.09E-02(+) & 2.0724±9.91E-02(+) & 2.3573±1.22E-01(+) & \textbf{3.4176±1.67E-02} \\
          & (5,10) & 1.8020±5.88E-02(+) & 1.1554±3.26E-02(+) & \textbf{3.3433±5.53E-02(-)} & 2.2123±9.47E-02(+) & 2.4650±1.04E-01(+) & 3.2589±2.68E-02 \\
          & (10,5) & 0.8896±1.03E-01(+) & 1.1361±4.98E-02(+) & 3.0427±6.01E-02(+) & 1.1253±1.08E-01(+) & 1.6316±2.34E-01(+) & \textbf{3.1048±2.45E-02} \\
    \midrule
    \multirow{3}[2]{*}{DF5} & (10,10) & 0.3326±5.39E-03(+) & 0.1216±4.12E-02(+) & 0.4425±7.35E-03(+) & 0.3338±4.36E-03(+) & 0.2746±3.71E-02(+) & \textbf{0.4624±3.06E-03} \\
          & (5,10) & 0.3277±4.31E-03(+) & 0.1256±2.03E-02(+) & \textbf{0.4331±9.40E-03(=)} & 0.3380±6.29E-03(+) & 0.3004±3.20E-02(+) & 0.4254±5.15E-03 \\
          & (10,5) & 0.2310±3.23E-03(+) & 0.1071±3.20E-02(+) & 0.3693±2.18E-02(+) & 0.2406±7.33E-03(+) & 0.2404±4.68E-03(+) & \textbf{0.3854±1.99E-02} \\
    \midrule
    \multirow{3}[2]{*}{DF6} & (10,10) & 0.0015±2.04E-03(+) & 0.0259±1.37E-02(+) & 0.2918±2.97E-02(=) & 0.0015±1.53E-03(+) & 0.0302±1.75E-02(=) & \textbf{0.3018±5.42E-02} \\
          & (5,10) & 0.0014±2.71E-03(+) & 0.0535±3.26E-02(+) & 0.0780±4.01E-02(+) & 0.0065±6.29E-03(+) & 0.0424±1.63E-02(+) & \textbf{0.1656±3.43E-02} \\
          & (10,5) & 0.0120±2.87E-03(+) & 0.0144±9.89E-03(+) & 0.0118±1.69E-02(+) & 0.0113±5.78E-03(+) & 0.0158±8.43E-03(+) & \textbf{0.0209±3.30E-02} \\
    \midrule
    \multirow{3}[2]{*}{DF7} & (10,10) & 0.0452±2.07E-02(+) & 0.0043±1.04E-02(+) & 3.3767±5.00E-02(+) & 0.2116±1.38E-01(+) & 1.3009±3.61E-01(+) & \textbf{3.4280±9.22E-02} \\
          & (5,10) & 0.1130±1.20E-01(+) & 0.6582±4.13E-01(+) & 2.5602±2.97E-01(+) & 0.3686±2.29E-01(+) & 1.5217±2.75E-01(+) & \textbf{3.0019±5.85E-01} \\
          & (10,5) & 0.0232±2.11E-02(+) & 0.2107±7.40E-01(+) & 0.9028±3.12E-01(+) & 0.0396±1.94E-02(+) & 1.2744±1.59E-01(+) & \textbf{1.7112±2.50E-01} \\
    \midrule
    \multirow{3}[2]{*}{DF8} & (10,10) & 51.818±4.83E-02(+) & 49.071±3.56E-02(+) & 52.378±5.24E-02(=) & 51.932±2.96E-02(+) & 52.354±3.60E-02(=) & \textbf{52.403±4.54E-02} \\
          & (5,10) & 53.590±2.01E-02(=) & 50.523±3.06E-02(+) & 53.970±1.96E-02(=) & 53.642±6.84E-02(=) & 53.858±3.74E-02(=) & \textbf{54.013±2.25E-02} \\
          & (10,5) & 51.412±3.38E-02(+) & 48.976±4.09E-02(+) & 52.126±7.43E-02(=) & 51.643±8.14E-02(+) & 52.197±5.06E-02(=) & \textbf{52.272±5.08E-02} \\
    \midrule
    \multirow{3}[2]{*}{DF9} & (10,10) & 15.630±3.13E-01(+) & 11.739±4.70E-01(+) & 16.229±4.54E-01(+) & 12.032±4.57E-01(+) & 14.140±3.56E-01(+) & \textbf{19.068±1.18E-01} \\
          & (5,10) & 14.526±2.20E-01(+) & 3.9067±1.19E+00(+) & 9.7689±6.89E-01(+) & 6.6475±1.22E-01(+) & 10.454±2.91E-01(+) & \textbf{17.069±8.85E-02} \\
          & (10,5) & 14.287±2.84E-01(+) & 11.941±5.46E-01(+) & 13.500±9.43E-01(+) & 10.162±4.12E-01(+) & 12.955±6.08E-01(+) & \textbf{17.707±1.74E-01} \\
    \midrule
    \multirow{3}[2]{*}{DF10} & (10,10) & 0.6101±3.79E-03(+) & 0.6164±2.98E-03(+) & 0.6883±2.73E-03(+) & 0.5969±2.03E-03(+) & 0.7330±9.24E-03(+) & \textbf{0.7878±3.72E-03} \\
          & (5,10) & 0.7650±2.50E-03(+) & 0.7324±4.55E-03(+) & 0.7945±3.60E-03(+) & 0.7205±4.80E-03(+) & 0.8304±4.60E-03(+) & \textbf{0.9064±7.19E-03} \\
          & (10,5) & 0.5517±2.66E-03(+) & 0.5432±5.16E-03(+) & 0.5892±4.84E-03(+) & 0.5338±3.23E-03(+) & 0.6566±8.06E-03(+) & \textbf{0.7308±4.11E-03} \\
    \midrule
    \multirow{3}[2]{*}{DF11} & (10,10) & 0.5900±1.37E-02(+) & 0.6462±1.07E-02(+) & 0.8824±5.66E-03(+) & 0.5537±1.30E-02(+) & 0.4029±4.69E-02(+) & \textbf{0.9695±3.03E-03} \\
          & (5,10) & 0.5872±8.29E-03(+) & 0.8023±1.44E-02(+) & \textbf{1.0555±2.27E-03(+)} & 0.6597±2.19E-02(+) & 0.4213±2.71E-02(+) & 0.9566±3.65E-03 \\
          & (10,5) & 0.3371±4.90E-03(+) & 0.4272±7.20E-03(+) & 0.7985±6.88E-03(+) & 0.3152±8.32E-03(+) & 0.3701±2.86E-02(+) & \textbf{0.9086±6.90E-03} \\
    \midrule
    \multirow{3}[2]{*}{DF12} & (10,10) & 9.8341±3.31E-03(+) & 8.3648±3.05E-02(+) & 9.8572±4.91E-03(=) & 9.8319±2.22E-03(+) & \textbf{9.8602±3.89E-02(=)} & 9.8592±5.61E-04 \\
          & (5,10) & 8.7664±7.23E-02(+) & 7.7746±2.01E-02(+) & 9.2294±2.75E-03(=) & 8.7230±5.27E-02(+) & \textbf{9.2302±2.45E-02(=)} & 9.2296±1.77E-04 \\
          & (10,5) & 9.6966±2.02E-02(+) & 8.0002±6.94E-02(+) & 9.8462±5.61E-03(=) & 9.6895±2.18E-02(+) & \textbf{9.8597±2.17E-02(=)} & 9.8532±3.18E-03 \\
    \midrule
    \multirow{3}[2]{*}{DF13} & (10,10) & 1.9718±3.09E-02(+) & 1.9045±6.61E-02(+) & 2.2321±2.67E-02(+) & 1.9371±2.35E-02(+) & 1.7975±5.13E-02(+) & \textbf{2.4104±3.66E-02} \\
          & (5,10) & 2.0124±1.41E-02(+) & 2.0079±4.28E-02(+) & 2.2706±1.47E-02(+) & 2.0188±1.16E-02(+) & 1.9317±4.53E-02(+) & \textbf{2.4010±2.16E-02} \\
          & (10,5) & 1.6846±1.36E-02(+) & 1.4227±1.51E-02(+) & 1.9582±4.04E-02(+) & 1.6584±1.15E-02(+) & 1.2634±4.19E-02(+) & \textbf{2.1315±3.37E-02} \\
    \midrule
    \multirow{3}[2]{*}{DF14} & (10,10) & 0.2796±9.17E-03(+) & 0.2833±6.96E-03(+) & 0.3623±1.48E-03(=) & 0.2806±5.31E-03(+) & 0.3211±1.40E-02(+) & \textbf{0.3722±2.53E-03} \\
          & (5,10) & 0.2760±4.31E-03(+) & 0.2824±5.36E-03(+) & 0.3475±3.10E-03(+) & 0.2788±6.19E-03(+) & 0.3255±5.99E-03(+) & \textbf{0.3608±7.40E-03} \\
          & (10,5) & 0.2130±4.25E-03(+) & 0.2398±2.79E-03(+) & 0.3190±7.60E-03(+) & 0.2106±4.29E-03(+) & 0.1947±3.05E-02(+) & \textbf{0.3442±6.72E-03} \\
    \bottomrule
    \end{tabular}
 \begin{tablenotes}
        \footnotesize
        \item (+), (=) and (-) indicate that ISVM-RM-MEDA performs significantly better or equivalently or worse than the compared algorithms, respectively.
      \end{tablenotes}
  \end{threeparttable}}
\end{table*}

The MIGD values indicate the comparatively convergence of solutions obtained by ISVM-RM-MEDA, and the MHV values indicate the superior diversity and distribution. In other words, the proposed ISVM-based prediction model can effectively explore the linear or nonlinear correlations between POS obtained at different environments, thereby generate promising populations for varying environments.

\subsection{Ablation Study}

In the proposed algorithm, two key procedures are implemented: SMOTE-based sampling strategy and ISVM-based prediction strategy. The results of comparison study have shown that the combination of the two strategies can significantly improve the quality of the predicted population. However, the role that each strategy plays in solving DMOPs remains unclear. To verify the effectiveness of these two strategies, we carry out an ablation experiment.
We modify the sampling strategy and propose two variants, in which the oversampling rate $r$ of POSMOTE is set to 0 and 3, and $r$ = 0 indicates that the sampling strategy is switched off. The two variants are denoted as ISVM$_{r0}$-RM-MEDA and ISVM$_{r3}$-RM-MEDA, respectively. In the exploration of the prediction strategy, we keep $r$ = 5 and deactivate the online update mechanism of ISVM. In other words, every time the environment changes, a new SVM is constructed based on the POS obtained from the former environment. The third variant is denoted as SVM-RM-MEDA.

Three variants were tested with the same parameters as the comparison experiment and the statistical results of MIGD averaged over three pairs of ($n_{t}$,$\tau_{t}$) configurations for each problem are shown in Table~\ref{table:ablation}. Consistent with the results in Table~\ref{table:MIGD comparision}, the average MIGD values obtained by ISVM-RM-MEDA are significantly better than those of DA-RM-MEDA on most problems with 30\%-60\% improvement. The only exceptions are DF8 and DF12, which are also challenges for other comparison algorithms.

ISVM$_{r0}$-RM-MEDA performs worse than DA-RM-MEDA on 6 out of 14 problems, and only a tiny improvement was observed on the rest of problems.
Invalidation of ISVM$_{r0}$-RM-MEDA can be attributed to the limited number of POS training samples, which is insufficient to build an accurate classifier.
With the samples tripled by the POSMOTE sampling strategy, ISVM$_{r3}$-RM-MEDA achieves significantly better MIGD results than ISVM$_{r0}$-RM-MEDA.
It is clear that inefficient predictions on DF1-DF3 and DF7 are corrected and the MIGD values on most problems are reduced. The quality of the solution sets obtained by ISVM-based algorithms is improved with the increase of the oversampling rate ($r$ = 0, 3, 5), which demonstrates the necessity and advantage of proposed sampling strategy.

SVM-RM-MEDA with the oversampling rate of 5 shows effective predictions.
But compared with ISVM-RM-MEDA, SVM-RM-MEDA still has room for improvement.
The superiority of ISVM-RM-MEDA over SVM-RM-MEDA can be attributed to the incremental learning mechanism, which incorporate more general features from samples in previous environments rather than only utilize the information from adjacent moment.

\begin{table}[htbp]
  \centering
  \caption{MIGD AVERAGED OVER THREE DYNAMIC CONFIGURATIONS OBTAINED BY ISVM-RM-MRDA AND ITS VARIANTS}
\label{table:ablation}
\resizebox{\linewidth}{2.4cm}{
\begin{threeparttable}
\begin{tabular}{m{2em}<{\centering}m{5em}<{\centering}m{5em}<{\centering}m{5em}<{\centering}m{5em}<{\centering}m{5em}<{\centering}}
    \toprule

    Prob & {\ \ DA-\newline{}RM-MEDA} & {ISVM$_{r0}$-\newline{}RM-MEDA} & {ISVM$_{r3}$-RM-MEDA} & {\ SVM-\newline{}RM-MEDA} & {\ ISVM-\newline{}RM-MEDA} \\      
\midrule
    DF1   & 0.2212 & 0.2517(-) & 0.1774(19.8) & 0.1090(50.7) & 0.0871(60.6) \\
    DF2   & 0.1434 & 0.1605(-) & 0.1125(21.5) & 0.0787(45.1) & 0.0679(52.6) \\
    DF3   & 0.4588 & 0.5234(-) & 0.3718(19.0) & 0.2698(41.2) & 0.2309(49.7) \\
    DF4   & 1.5398 & 1.4585(5.3) & 1.3069(15.1) & 1.1064(28.1) & 0.9935(35.5) \\
    DF5   & 1.7366 & 1.6074(7.4) & 1.4552(16.2) & 1.3175(24.1) & 1.0924(37.1) \\
    DF6   & 8.6197 & 8.2571(4.2) & 8.0207(6.9) & 6.0058(30.3) & 5.0582(41.3) \\
    DF7   & 7.3851 & 7.4237(-) & 6.9706(5.6) & 5.8792(20.4) & 4.6844(36.6) \\
    DF8   & 0.7116 & 0.8249(-) & 0.7743(-) & 0.7697(-) & 0.7693(-) \\
    DF9   & 2.3775 & 2.3015(3.2) & 2.0593(13.4) & 1.8020(24.2) & 1.6047(32.5) \\
    DF10  & 0.1732 & 0.1611(7.0) & 0.1374(20.7) & 0.1236(28.6) & 0.1125(35.0) \\
    DF11  & 0.2424 & 0.2243(7.5) & 0.1694(30.1) & 0.1209(50.1) & 0.0893(63.2) \\
    DF12  & 0.6825 & 1.1603(-) & 1.1195(-) & 1.1638(-) & 1.1594(-) \\
    DF13  & 1.7704 & 1.5978(9.7) & 1.4464(18.3) & 1.3293(24.9) & 1.2280(30.6) \\
    DF14  & 1.2048 & 1.1334(5.9) & 0.9247(23.2) & 0.8651(28.2) & 0.7406(38.5) \\
    \bottomrule
    \end{tabular}
\begin{tablenotes}
        \footnotesize
        \item The values in parentheses indicate the percentage improvement versus DA-RM-MEDA, and (-) indicates no improvement.
      \end{tablenotes}
  \end{threeparttable}}
\end{table}

\subsection{Adaptation Study}

In the experiments above, RM-MEDA is employed to optimize the population in each iteration. To further investigate if ISVM-DMOEA relies on a specific static multiobjective optimization algorithm, we choose two other popular static algorithms as the optimizer. The first one is NSGA-II \cite{deb2002fast}, a genetic algorithm that uses non-dominant sorting and crowding distance to select dominant individuals derived from cross mutation. The second one is the multiple objective particle swarm optimization algorithm \cite{coello2004handling}, abbreviated as MOPSO, in which the flight direction of particles determined by Pareto dominance is regarded as a guide for solution search. 

\begin{table}[htbp]
  \centering
  \caption{MIGD AVERAGED OVER THREE DYNAMIC CONFIGURATIONS OBTAINED BY DA-NSGA-II, ISVM-NSGA-II, DA-MOPSO AND ISVM-MOPSO}
\label{table:adaptation}
\resizebox{\linewidth}{2.45cm}{
\begin{threeparttable} 
    \begin{tabular}{ccccc}
    \toprule
    Prob & DA-NSGA-II & ISVM-NSGA-II & DA-MOPSO & ISVM-MOPSO \\
    \midrule
    DF1   & 0.5660 & 0.4381(22.6) & 0.1315 & 0.0488(62.9) \\
    DF2   & 0.4855 & 0.3132(35.5) & 0.0887 & 0.0282(68.2) \\
    DF3   & 0.6347 & 0.3744(41.0) & 0.2008 & 0.1294(35.6) \\
    DF4   & 1.9088 & 1.2459(34.7) & 1.4380 & 0.8301(42.3) \\
    DF5   & 2.0731 & 1.5564(24.9) & 0.3078 & 0.1380(55.2) \\
    DF6   & 12.650 & 12.299(2.8) & 4.4071 & 0.3611(91.8) \\
    DF7   & 12.281 & 10.224(16.7) & 1.6328 & 0.5933(63.7) \\
    DF8   & 0.6447 & 0.7094(-) & 0.6074 & 0.6642(-) \\
    DF9   & 2.4538 & 2.2416(8.6) & 0.8732 & 0.8358(4.3) \\
    DF10  & 0.2745 & 0.1692(38.4) & 0.1928 & 0.1262(34.5) \\
    DF11  & 0.4788 & 0.2822(41.1) & 0.2289 & 0.1259(45.0) \\
    DF12  & 0.6737 & 0.8052(-) & 0.4208 & 0.7478(-) \\
    DF13  & 2.1968 & 1.8398(16.3) & 0.2710 & 0.2552(5.8) \\
    DF14  & 1.4877 & 1.3651(8.2) & 0.4496 & 0.3981(11.5) \\
    \bottomrule
    \end{tabular}
\begin{tablenotes}
        \footnotesize
        \item The values in parentheses indicate the percentage improvement of ISVM-NSGA-II versus DA-NSGA-II and that of ISVM-MOPSO versus DA-MOPSO, respectively. (-) indicates no improvement.
      \end{tablenotes}
  \end{threeparttable}}
  \label{tab:addlabel}
\end{table}

Similar to the operation on RM-MEDA, these two methods are embedded in ISVM-DMOEA and also modified to adapt to dynamic change. The modified algorithms are called DA-NSGA-II, ISVM-NSGA-II, DA-MOPSO and ISVM-MOPSO, respectively. The results of average MIGD obtained by the four algorithms on DF1-DF14 are shown in Table~\ref{table:adaptation}. 

Due to different static optimization strategies, the average MIGD values obtained by three kinds of algorithms are quite different. However, it is clear from the table that ISVM-based algorithms for NSGA-II and MOPSO significantly surpass the randomly reinitialized algorithms, which is also observed in RM-MEDA. 
ISVM-MOPSO greatly improves the performance of DA-MOPSO and even reduces the average MIGD values by more than 60\% on four problems, because the high-quality initial population accelerates the convergence of particle swarm to the optimal solutions.

The strong adaptability and outstanding performance on RM-MEDA, NSGA-II and MOPSO demonstrate that ISVM-DMOEA is a versatile algorithm that can effectively cooperate with various static optimization algorithms.

\section{Conclusion}
\label{sec:Conclusion}

Prediction-based models are promising in solving dynamic multiobjective optimization problems, which estimate the optimal solutions by taking advantage of existing information.
An assumption is put forward that predictable general patterns can be detected from some implicit correlations between the optimal solutions obtained in changing environments. 
In this paper, we propose an online prediction model based on incremental support vector machine (ISVM) to utilize these patterns.
The parameters of ISVM are updated online to accommodate new samples and extract the linear or nonlinear (more common in practice) correlations between previous POS.
When the environment changes, an evolving ISVM classifier can estimate whether a solution has the characteristics of being an optimal solution and guide the prediction of the initial population.
To avoid low prediction accuracy caused by the limited number of POS samples, a sampling strategy based on SMOTE is implemented in advance.

The experimental results demonstrate that the proposed algorithm achieves competitive performance on DMOPs. 
Besides, ablation and adaptation study indicate the versatility of ISVM-DMOEA and the advantage of sub-strategies.
However, solving problems with low correlation between solutions at different environments and problems with strange distributions such as POF holes remains a challenge.
For future work, we would like to combine ISVM-DMOEA with transfer learning to improve the applicability of the algorithm on more instances.
In addition, we plan to introduce few-shot learning and sample quality pre-evaluation, which can enhance the accuracy of prediction.
Furthermore, kinds of machine learning methods are expected to be integrated into evolutionary algorithm for solving real-world problems.

\ifCLASSOPTIONcaptionsoff
  \newpage
\fi

\bibliographystyle{IEEEtran}

\bibliography{ISVMDMOEAbib}

\end{document}